\documentclass{article}

\usepackage[dblblindworkshop, final]{neurips_2026}
\workshoptitle{Global South AI @ NeurIPS 2026: Rethinking AI for and from the Global South}

\usepackage[utf8]{inputenc}
\usepackage[T1]{fontenc}
\usepackage{hyperref}
\hypersetup{hidelinks}
\usepackage{url}
\usepackage{booktabs}
\usepackage{amsfonts}
\usepackage{amsmath}
\usepackage{microtype}
\usepackage{xcolor}
\usepackage{graphicx}
\usepackage{multirow}
\usepackage{tikz}
\usetikzlibrary{arrows.meta,positioning,calc}

\title{One Threshold Does Not Fit All Languages:\\Language-Conditional Deferral for Reliable and Efficient Low-Resource Text Classification}

\author{%
  Bhanu Prakash Vangala \\
  University of Missouri \\
  \texttt{bv3hz@missouri.edu} \And
  Vangala Navya \\
  Amar Bio Tech Pvt Ltd \\
  \texttt{vangalanavya.8@gmail.com}
}

\begin{document}

\maketitle

\begin{abstract}
In the Global South, the lower-income countries of Africa, Asia, and Latin America where most of the world's languages are spoken, a deployed text classifier usually runs on ordinary CPUs, serves many languages with a single model, has few labeled examples in any of them, and relies on people to catch its mistakes. Such a system is only useful if it can promise how often it will be wrong: at most a fixed fraction of the labels it assigns on its own may be incorrect, and everything else must go to a person. Split conformal prediction delivers this promise through a single confidence threshold, normally estimated on validation data pooled across languages. We ask whether the promise reaches every language, and it does not. On MasakhaNEWS (16 African languages) and AfriSenti (12 languages plus two never seen in training), a pooled threshold meets the 90\% target on average but covers Somali at 77.5\%, Tigrinya at 83.7\%, and the two unseen languages at 77.5\% and 81.2\%. Estimating one threshold per language brings every language to between 89.1\% and 91.0\% without retraining, and it shows how unequal the cost of the promise is: keeping it means sending 43\% of Somali news and over 80\% of Amharic and Xitsonga tweets to a person, against under 8\% of Nigerian Pidgin news. One or two hundred labels per language are enough and the models train in minutes on one CPU core, so the fix is affordable: calibrate, report, and budget human review one language at a time.
\end{abstract}

\section{Introduction}

The Global South is the usual name for the lower-income countries of Africa, Latin America, and much of Asia and Oceania. It is home to most of the world's population and most of its roughly 7,000 languages, and almost all of those languages have little or no labeled data~\citep{joshi2020state}. When a ministry, a hospital network, or a newsroom in this setting deploys a text classifier, the conditions differ from those most NLP research assumes. There is no GPU, so the model must train and run on ordinary CPUs. There is no budget for one model per language, so a single model serves all of them. Labeled data exists in the hundreds of examples rather than the hundreds of thousands, and it is produced by communities of native speakers~\citep{nekoto2020participatory}. And because the classifier sorts things that matter, such as citizen complaints, health-hotline messages, or news that reaches an editor, a person reviews whatever the model is not sure about.

In this setting the property a deployer cares about is not average accuracy but how often the system will be wrong when it acts on its own. We call the corresponding promise a \emph{reliability contract}: among all inputs, at most a fraction $\alpha$ receive a wrong automatic label, and the rest are handed to a person or to a larger, more expensive model. Selective prediction with a reject option is the classical way to obtain such a contract~\citep{chow1970optimum,geifman2017selective}, and split conformal prediction turns it into a guarantee that holds for any model, using nothing more than a held-out calibration set and a single confidence threshold~\citep{vovk2005algorithmic,angelopoulos2023gentle}.

The problem this paper studies is where that threshold comes from. The obvious engineering choice is to estimate one threshold on validation data pooled across all languages. The guarantee this produces is only marginal: it holds on average over the mix of languages in the calibration data and says nothing about any single language~\citep{vovk2012conditional,romano2020malice}. \citet{jones2021selective} showed that selective classification can widen gaps between groups, and languages are groups of exactly this kind. The failure is also easy to miss, because the dashboard reports the aggregate: it reads 90\% coverage, and nobody checks Somali. So the question is concrete: does a single pooled threshold deliver the promised reliability to every language, and if not, what does it take, in labels and in human review, to deliver it?

We answer this on two benchmarks of African-language text classification built by African NLP communities~\citep{adelani2023masakhanews,muhammad2023afrisenti}, using a single CPU core, the regime in which low-resource languages are usually the first to lose out~\citep{ahia2021double,alabi2022adapting}. A pooled threshold satisfies the contract in aggregate while under-covering the lowest-resource languages by up to 12.5 points. Estimating one threshold per language, the Mondrian construction of \citet{vovk2012conditional} that underlies equalized coverage~\citep{romano2020malice,gibbs2023conditional}, repairs this without retraining anything. It also reveals what the contract costs in each language, and it needs only 100 to 200 labeled examples per language. Code, seeds, and results accompany the submission.

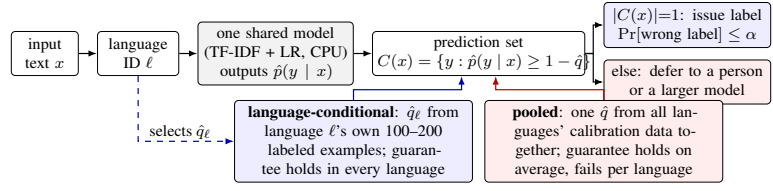
\begin{figure}[t]
  \centering
  \resizebox{0.72\linewidth}{!}{%
  \begin{tikzpicture}[
    font=\footnotesize,
    box/.style={draw, line width=0.5pt, rounded corners=2pt, align=center, inner sep=2.5pt, minimum height=2.2em},
    arr/.style={-{Latex[length=1.5mm]}, line width=0.6pt},
    node distance=3mm and 3.2mm]
    \node[box, text width=1.0cm] (x) {input\\text $x$};
    \node[box, right=of x, text width=1.25cm] (lid) {language\\ID $\ell$};
    \node[box, right=of lid, fill=gray!10, text width=2.55cm] (model) {one shared model\\(TF-IDF + LR, CPU)\\outputs $\hat p(y\mid x)$};
    \node[box, right=of model, text width=3.55cm] (set) {prediction set\\\mbox{$C(x)=\{y:\hat p(y\mid x)\ge 1-\hat q\}$}};
    \node[box, right=of set, yshift=5mm, fill=blue!7, text width=2.75cm] (auto) {$|C(x)|{=}1$: issue label\\\mbox{$\Pr[\text{wrong label}]\le\alpha$}};
    \node[box, right=of set, yshift=-5mm, fill=red!7, text width=2.75cm] (defer) {else: defer to a person\\or a larger model};
    \draw[arr] (x) -- (lid);
    \draw[arr] (lid) -- (model);
    \draw[arr] (model) -- (set);
    \draw[arr] (set.east) -- ++(1.4mm,0) |- (auto.west);
    \draw[arr] (set.east) -- ++(1.4mm,0) |- (defer.west);
    \node[box, fill=blue!7, text width=4.0cm, anchor=north east] (cond) at ($(set.south)+(-1mm,-4.2mm)$)
      {\textbf{language-conditional}: $\hat q_\ell$ from language $\ell$'s own 100--200 labeled examples; guarantee holds in every language};
    \node[box, fill=red!7, text width=4.0cm, anchor=north west] (pooled) at ($(set.south)+(1mm,-4.2mm)$)
      {\textbf{pooled}: one $\hat q$ from all languages' calibration data together; guarantee holds on average, fails per language};
    \draw[arr, blue!70!black] (cond.north) -- ++(0,1.8mm) -| ([xshift=-3mm]set.south);
    \draw[arr, red!70!black] (pooled.north) -- ++(0,1.8mm) -| ([xshift=3mm]set.south);
    \draw[arr, dashed, blue!70!black] (lid.south) |- (cond.west) node[pos=0.72, above, inner sep=1pt, black] {selects $\hat q_\ell$};
  \end{tikzpicture}}
  \caption{The deferral pipeline. One cheap model serves every language; a conformal threshold turns its probabilities into a prediction set, and only sets with exactly one label become automatic labels. The two policies differ only in where the threshold comes from.}
  \label{fig:pipeline}
\end{figure}

\section{Reliability contracts from a single cheap model}
\label{sec:method}

\paragraph{Problem statement.} A deployment has a fixed classifier that outputs probabilities $\hat p(y \mid x)$ for inputs from $L$ languages, and a target error rate $\alpha$ for automatic decisions. For every input it must decide whether to issue the model's label or to defer the input to a person. We want a deferral rule with four properties: in every language, at most a fraction $\alpha$ of inputs receive a wrong automatic label; as little traffic as possible is deferred subject to that guarantee; no retraining is needed and only a small number of labeled examples per language; and the deferral rate of each language, which is its human-review cost, is visible to the deployer. Figure~\ref{fig:pipeline} shows the pipeline we study.

\paragraph{Model.} For each task we train a single multilingual model on the union of all languages' training sets: TF-IDF weights over character $n$-grams (lengths 2 to 5, within word boundaries) and word unigrams and bigrams, capped at 450k features, followed by multinomial logistic regression. The calibration layer sees nothing but the model's probabilities, so everything that follows applies equally to a fine-tuned encoder or to a model served through an API.

\paragraph{Conformal selective prediction.} We use the least-ambiguous set-valued classifier of \citet{sadinle2019least} in its split conformal form. Given a calibration set $\{(x_i, y_i)\}_{i=1}^{n}$, we compute a score $s_i = 1 - \hat p(y_i \mid x_i)$ for each example and take the finite-sample quantile
\begin{equation}
\hat q = s_{(k)}, \qquad k = \lceil (n+1)(1-\alpha) \rceil ,
\label{eq:quantile}
\end{equation}
that is, the $k$-th smallest score. For a new input the prediction set is $C(x) = \{ y : \hat p(y \mid x) \ge 1 - \hat q \}$, and if calibration and test inputs are exchangeable, the true label lies in this set with probability at least $1-\alpha$~\citep{vovk2005algorithmic,lei2018distribution}. The deployment rule is simple: if $C(x)$ contains exactly one label, the system issues that label; otherwise the input is deferred. A wrong automatic label can only occur when a singleton set misses the true label, which is a miscoverage event, so the probability of a wrong automatic label over all traffic is at most $\alpha$. This inequality is the reliability contract; we use $\alpha = 0.10$ throughout.

\paragraph{Pooled versus language-conditional calibration.} The \emph{pooled} policy computes a single $\hat q$ from the calibration data of all languages together. The \emph{language-conditional} policy computes a separate $\hat q_\ell$ for each language $\ell$ from that language's calibration data alone; this is the Mondrian construction of \citet{vovk2012conditional} with language as the taxonomy, and it guarantees coverage of at least $1-\alpha$ within every language. Neither policy changes the trained model (unlike learning to defer~\citep{madras2018predict,mozannar2020consistent}); the conditional policy additionally needs the language of each input, which a multilingual deployment already needs for routing, and a few labeled examples per language.

\section{Experiments}
\label{sec:experiments}

\paragraph{Data and protocol.} MasakhaNEWS~\citep{adelani2023masakhanews} is news topic classification with seven categories in 16 African languages; training sets range from 608 articles (Lingala) to 3,309 (English), and we use the headline plus the first 1,000 characters of the body. AfriSenti~\citep{muhammad2023afrisenti,muhammad2023semeval} is three-way tweet sentiment in 12 languages with training data plus Oromo and Tigrinya, which have development and test data only and therefore act as languages the model has never seen. Conformal validity assumes that calibration and evaluation data are exchangeable, so for every language we pool the official development and test splits and, for each of 30 random seeds, divide them into a calibration half and an evaluation half; all numbers are means (and standard deviations) over these seeds. The pooled threshold uses the calibration halves of the trained languages only, and the calibration-size ablation draws $n \in \{20, 50, 100, 200\}$ examples from a language's calibration half.

\paragraph{Efficiency and base quality.} On a single CPU core, training takes 3.4 minutes for MasakhaNEWS and 1.3 minutes for AfriSenti, and the models occupy 40 MB and 25 MB. Accuracy on the official test splits averages 87.6\% on MasakhaNEWS (macro-F1 82.4, about ten points below the best fully supervised result of \citet{adelani2023masakhanews}) and 62.9\% on AfriSenti (macro-F1 57.4). The model is deliberately cheap; the point is that a correctly calibrated contract makes it safe to deploy.

\begin{figure}[t]
  \centering
  \includegraphics[width=0.78\linewidth]{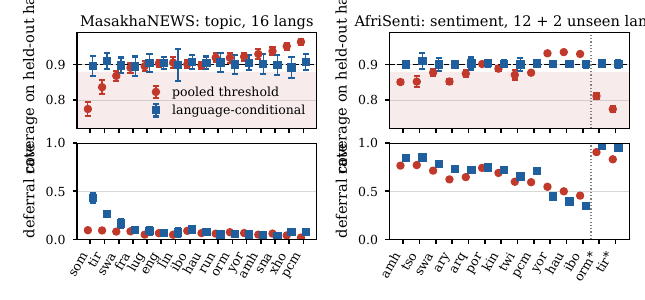}
  \caption{Coverage (top) and deferral rate (bottom) per language at a 0.90 target (dashed), mean $\pm$ s.d.\ over 30 random splits. Trained languages are ordered by base accuracy; starred languages never appeared in training. A pooled threshold (red) over-covers well-resourced languages and under-covers hard ones (shaded region); a per-language threshold (blue) meets the target everywhere, at a very different deferral cost per language.}
  \label{fig:main}
\end{figure}

\subsection{A pooled threshold keeps the contract on average and breaks it per language}

Figure~\ref{fig:main} (top) contains the central result. With the pooled policy, aggregate coverage is $0.902 \pm 0.006$ on MasakhaNEWS and $0.901 \pm 0.003$ on AfriSenti, exactly the promised 0.90, but language by language the picture reverses. On MasakhaNEWS, Somali is covered at $0.775 \pm 0.019$, Tigrinya at $0.837 \pm 0.019$, and Swahili at $0.868 \pm 0.015$, while Nigerian Pidgin, isiXhosa, and chiShona reach $0.964$, $0.951$, and $0.938$. On AfriSenti, seven of the twelve trained languages fall more than two points below the target, while Hausa, Igbo, and Yoruba are over-covered at $0.93$. The reason is that the pooled quantile in Equation~\ref{eq:quantile} is dominated by the languages with the most calibration examples and the most confident predictions, so it is too permissive exactly where the model is least sure. Concretely, a Somali article that the pooled system labels on its own is correct $79.5\%$ of the time, not the $90\%$ that the aggregate figure suggests.

The language-conditional policy removes this disparity. Every language lands between $0.891$ and $0.910$ on MasakhaNEWS and between $0.900$ and $0.910$ on AfriSenti, aggregate coverage stays at $0.902$, and automatic Somali labels are now $94.4\%$ correct. In numbers: three MasakhaNEWS and seven AfriSenti languages sit more than two points below target under the pooled threshold and none under the conditional one, whose overall deferral rises from $6.6\%$ to $10.1\%$ on MasakhaNEWS but falls from $59.4\%$ to $58.4\%$ on AfriSenti as review effort moves to the languages that need it.

\subsection{The price of reliability is not the same in every language}

Meeting the contract in every language makes its cost explicit (Figure~\ref{fig:main}, bottom). To label Somali news automatically at 90\% reliability, the conditional policy has to defer $42.9\%$ of Somali traffic, against $9.5\%$ under the pooled threshold. Tigrinya needs $26.4\%$, Swahili $16.5\%$, and Amharic, Yoruba, and chiShona under $5\%$. On AfriSenti the spread is wider still: $84.7\%$ of Amharic and $85.1\%$ of Xitsonga tweets are deferred, compared with $34.6\%$ of Igbo tweets. The deferral rate closely follows base accuracy (Spearman $\rho = -0.70$ on MasakhaNEWS, $-0.97$ on AfriSenti). These rates are the human-review budget a deployment has to plan for in each language; a pooled threshold hides them and spends the budget on the languages that need it least.

\paragraph{Languages the model has never seen.} With the pooled threshold, the system labels $9.3\%$ of Oromo and $16.8\%$ of Tigrinya tweets on its own, and those labels are correct only $33.8\%$ and $37.6\%$ of the time, which is chance level for three classes: a pooled threshold lets the model act confidently on a language it cannot read. A language-specific threshold estimated from a few hundred examples defers $96.8\%$ and $95.0\%$ of that traffic and restores $0.90$ coverage.

\subsection{How many labels does a language need, and where must they come from?}

Per-language calibration needs labeled examples in each language, the scarcest resource in the Global South, but the requirement turns out to be modest. Averaged over languages, coverage is $0.904$ with only 20 calibration examples and between $0.897$ and $0.903$ with 50 or more: the guarantee is unbiased even for tiny calibration sets. What shrinks with $n$ is the variability of the coverage any one deployment experiences: the standard deviation across seeds falls from $0.06$ at $n = 20$ to about $0.03$ at $n = 100$ and $0.02$ at $n = 200$. Smaller calibration sets also defer more ($17.5\%$ at $n = 20$ against $10.9\%$ with the full half on MasakhaNEWS); in practice, 100 to 200 labels per language are enough.

The labels also have to come from the right distribution. Calibrating on the official development splits and evaluating on the official test splits leaves MasakhaNEWS unchanged but drops AfriSenti's aggregate coverage to $0.879$, with per-language thresholds under-covering Moroccan Arabic ($0.798$), Algerian Arabic ($0.848$), and Nigerian Pidgin ($0.848$); random re-splits of the same data cover all three at $0.90$. Those released splits are not exchangeable, and a contract certified on a benchmark holds in deployment only if the calibration data come from the traffic the system will actually see.

\section{Recommendations and limitations}

Our results support four practices for anyone who deploys a multilingual classifier with a human fallback. Calibrate the deferral threshold per language, never on pooled data. Report coverage and deferral per language, because the aggregate hides the gap. Budget human review per language from the measured deferral rates, treating a high rate as a reason to collect training data rather than to lower the bar. Two limits: a fine-tuned encoder would defer less than our cheap model, though the pooled-versus-conditional gap belongs to the calibration layer, not the model; and the conditional policy relies on language identification, itself weaker for low-resource languages. One threshold does not fit all languages, and the languages it fits worst are the ones this workshop exists for.

\begin{ack}
Hidden in the anonymized submission.
\end{ack}

{\small
\setlength{\bibsep}{2.5pt}
\bibliographystyle{plainnat}
\bibliography{references}
}

\end{document}